\documentclass[runninheads]{llncs}
\usepackage{bm}
\usepackage[utf8]{inputenc}
\usepackage{xurl}
\usepackage{comment}
\usepackage{amssymb}
\usepackage{amsfonts}
\usepackage{graphicx}
\usepackage{url}
\usepackage{lstsemantic}
\usepackage{lstlangmizar}
\usepackage{booktabs}
\usepackage{footnoterange}
\usepackage{todonotes}
\usepackage{cite}
\usepackage{siunitx}

\usepackage{hyperref}

\usepackage{doc}

\usepackage{paralist}

\newcommand*{\ModelSmallLgb}{\ensuremath{\mathcal{D}_\textsf{small}}}
\newcommand*{\ModelLargeLgb}{\ensuremath{\mathcal{D}_\textsf{large}}}
\newcommand*{\ModelSmallGnn}{\ensuremath{\mathcal{G}_\textsf{small}}}
\newcommand*{\ModelLargeGnn}{\ensuremath{\mathcal{G}_\textsf{large}}}
\renewcommand*{\S}{\mathcal{S}}

\newcommand*{\ModelParS}{\ensuremath{\mathcal{P}^\textsf{given}}}
\newcommand*{\ModelParP}{\ensuremath{\mathcal{P}^\textsf{proof}}}
\newcommand*{\ModelParM}{\ensuremath{\mathcal{P}_\textsf{fuse}}}
\newcommand*{\ModelParC}{\ensuremath{\mathcal{P}_\textsf{cat}}}
\newcommand*{\ModelParMS}{\ensuremath{\mathcal{P}_\textsf{fuse}^\textsf{given}}}
\newcommand*{\ModelParMP}{\ensuremath{\mathcal{P}_\textsf{fuse}^\textsf{proof}}}

\title{Fast and Slow Enigmas and Parental Guidance
}

\author{
   Zarathustra A.~Goertzel\inst{1} %
   \and  Karel Chvalovsk\'y\inst{1}
   \and  Jan Jakub\r{u}v\inst{1,2}
   \and  Miroslav Ol\v{s}\'ak\inst{2}
   \and Josef Urban\inst{1} %
}

\institute{
   Czech Technical University in Prague, Prague, Czech Republic\\
\and
   University of Innsbruck, Austria
 }

\authorrunning{Goertzel}%

\titlerunning{Two-phase Enigma}

\makeatletter
\renewcommand\section{\@startsection{section}{1}{\z@}%
                       {-12\p@ \@plus -4\p@ \@minus -4\p@}%
                       {8\p@ \@plus 4\p@ \@minus 4\p@}%
                       {\normalfont\large\bfseries\boldmath
                        \rightskip=\z@ \@plus 8em\pretolerance=10000 }}
\makeatother

\begin{document}

\maketitle

\begin{abstract}
We describe several additions to the ENIGMA system that guides clause
selection in the E automated theorem prover. First, we significantly speed up its neural guidance by adding
server-based GPU evaluation. The second addition is motivated by fast weight-based rejection filters that are currently
used in systems like E and Prover9. Such systems can be made more
intelligent by instead training fast versions of ENIGMA that
implement more intelligent pre-filtering. This results in combinations of trainable fast and slow
thinking that improves over both the fast-only and slow-only methods.
The third addition is
based on "judging the children by their parents", i.e., possibly
rejecting an inference before it produces a clause. This is
motivated by standard evolutionary mechanisms, where there is always a
cost to producing all possible offsprings in the current population. 
This saves time by not evaluating all clauses by more expensive methods 
and provides a complementary view of the generated clauses.
The methods are evaluated on a large benchmark coming from the Mizar
Mathematical Library, showing good improvements over the state of the art.
\end{abstract}

\section{Introduction: The Fast and The Smart}
\label{sect:introduction}

Throughout the history of automated theorem proving, there have been
two very different approaches to strengthening automated theorem
provers (ATPs). The first one (\emph{the fast}) relies on
better engineering, such as improving the indexing for inference and
reduction rules and on optimized low-level implementations. The
gains achieved in this way can be quite
high~\cite{stickel1989path,McCune92a,Voronkov95,Hillenbrand03,Schulz12,DBLP:conf/tableaux/Kaliszyk15}.

The
second approach (\emph{the smart}) relies on advanced strategies and heuristics
for guiding the proof search.  This includes methods using extensive
previous knowledge, e.g., various kinds of \emph{symbolic}
machine learning, such as the \emph{hints} method in
Otter~\cite{Veroff96} and Prover9~\cite{KinyonVV13}, and its
\emph{watchlist}~\cite{DBLP:conf/cade/Ruhdorfer020} and
\emph{proofwatch}~\cite{DBLP:conf/itp/GoertzelJ0U18} variants
implemented in E~\cite{Schulz13,Schulz19}. With the recent advent of
\emph{statistical} machine learning (ML), a number of 
knowledge-based ATP-guiding methods have been %
created~\cite{JakubuvU17a,KaliszykUMO18,DBLP:conf/cade/ChvalovskyJ0U19,DBLP:conf/cade/JakubuvCOP0U20}. This
is done by compiling (extracting, compressing, generalizing) the
previous knowledge into statistical ML \emph{predictors} (models) that are then
used to predict the usefulness of inference steps in
the proof search. 

The \emph{smart} approaches, while potentially sophisticated and AI-motivated, may
incur prohibitively high costs in their prediction modules, in
particular when naively implemented~\cite{UrbanVS11,LoosISK17}. 
This can make them inferior in practice 
to faster alternative approaches, such as various kinds
of randomization~\cite{DBLP:conf/cade/RathsO08} and building of
portfolios of complementary fast
strategies~\cite{blistr,SchaferS15,DBLP:journals/aicom/JakubuvU18}.
This issue is getting increasingly important as deep learning (DL) is
used for ATP guidance, sometimes with large cloud-based DL-predictors
running on specialized hardware that hides the amount of resources
used. It also complicates rigorous comparisons in established ATP competitions such as
CASC/LTB~\cite{SutcliffeS06,SutcliffeU15}.

Another issue related to the use of expensive predictors can be
summarized as the \emph{explore-exploit tradeoff} introduced in
reinforcement learning research~\cite{gittins1979bandit}. In short, running an ATP guided by a
100-times slower predictor that is only slightly better (possibly due
to insufficient previous data for learning) will not only typically
solve fewer problems due to much more expensive backtracking but also
generate much less data for training the predictor in the next
iteration. Hence, given a global time limit allowing many
proving/learning iterations over a large set of related problems in a
realistic problem-solving setup such as CASC LTB, a faster predictor will in the same time generate
much more data to learn from. This in turn often leads to
better performance: a slightly weaker ML system trained on much more data
will often ultimately outperform a slightly stronger ML system trained on much less data.

\subsection{Contributions}
In this work we develop combinations of the fast(er) and smart(er)
approaches in the context of the learning-guided ENIGMA framework.
After giving a summary of ENIGMA in Section~\ref{enigma}, Section~\ref{sec:filtering} introduces our new methods.\footnote{The
  E and ENIGMA versions used in this paper can be found at \url{https://github.com/ai4reason/enigma-gpu-server}.}

First, Section~\ref{server} describes a
large increase in the speed of neural guidance in ENIGMA. We add an 
efficient server-based evaluation that uses dedicated GPUs instead of a
CPU. When using four commodity GPU cards, this speeds up the neural evaluation of the clauses about four times in real time. 

Section~\ref{gnn-and-gbdt} describes the second addition, motivated
by fast weight-based rejection filters used in systems such as E and Prover9. Such methods can be replaced
by training fast predictors that implement
more intelligent pre-filtering. In the context of ENIGMA, fast(er) is
easy to implement by variously parameterized predictors based on
gradient-boosted decision trees (GBDTs). Slow(er) models are in
those based on graph neural networks (GNNs). 

Section~\ref{parental} describes the third addition
based on "judging the children by their parents", i.e., possibly
rejecting an inference before it even produces a clause. This
grants the machine learning methods greater control 
of the proof search and saves time by not evaluating all clauses by more expensive methods, 
also providing a complementary view of the generated clauses.

In Section~\ref{setting} we describe the experimental setting and a
large evaluation corpus based on the Mizar Mathematical Library and
its MPTP translation. We also present our baseline methods there. 
Section~\ref{evaluation} evaluates the new methods and shows that
even in relatively low time limits the methods provide
good performance improvements over the previous versions of ENIGMA.

\section{Saturation Proving and Its Guidance by ENIGMA}
\label{enigma}

State-of-the-art automated theorem provers (ATP), such as E, Prover9, and
Vampire~\cite{Vampire}, are based on the saturation loop paradigm and the \emph{given clause
algorithm}~\cite{Overbeek:1974:NCA:321812.321814}.
The input problem, in first-order logic (FOF), is translated into a refutationally
equivalent set of clauses, and a search for contradiction is initiated.
The ATP maintains two sets of clauses: \emph{processed} (initially empty) and
\emph{unprocessed} (initially the input clauses).  
At each iteration, one unprocessed clause is selected (\emph{given}), and all of the 
possible inferences with all the processed clauses are generated 
(typically using resolution, paramodulation, etc.), extending the unprocessed clause set.
The selected clause is then moved to the processed clause set.
Hence the invariant holds that all the mutual inferences among the processed
clauses have been computed.

The selection of the ``right'' given clause is known to be %
vital for the success of the proof search.
The ENIGMA
system~\cite{JakubuvU17a,JakubuvU18,JakubuvU19,GoertzelJU19,DBLP:conf/cade/ChvalovskyJ0U19,DBLP:conf/cade/JakubuvCOP0U20}
applies various machine learning methods for given clause selection, learning
from a large number of previous successful proof searches.
The training data consists of clauses processed during a proof search, labeling
the clauses that appear in the discovered proof as \emph{positive}, and the other
(thus unnecessary) processed clauses as \emph{negative}.

The first ENIGMA~\cite{JakubuvU17a} used fast linear
classification~\cite{Fan:2008:LLL:1390681.1442794} with hand-crafted clause
\emph{features} based on symbol names, representing clauses by 
fixed-length numeric vectors.
Follow-up
versions~\cite{JakubuvU18,JakubuvU19,GoertzelJU19,DBLP:conf/cade/ChvalovskyJ0U19}
introduced context-based clause evaluation and
fast dimensionality reduction by feature hashing,
and employed Gradient Boosting
Decision Trees (GBDTs), implemented by the XGBoost and LightGBM systems~\cite{Chen:2016:XST:2939672.2939785,LightGBM}),
and Recursive Neural
Networks (implemented in PyTorch) as the underlying machine learning methods.

The latest version, ENIGMA Anonymous~\cite{DBLP:conf/cade/JakubuvCOP0U20}, 
abstracts from name-based clause representations and provides
the best results so far both with GBDTs and Graph Neural Networks
(GNNs)~\cite{AAB:2016tensorflow-2016}.
For GBDTs, clauses are again represented by fixed-length vectors based on
syntax trees and
anonymization is achieved by replacing
symbol names by their arities.
Our GNN~\cite{DBLP:conf/ecai/OlsakKU20} represents clauses by variable-length numeric tensors
encapsulating syntax trees as graph
structures with symbol names omitted.
ENIGMA-GNN evaluates new clauses jointly in larger batches (\emph{queries}) and with respect
to a large number of already selected clauses (\emph{context}). The GNN
predicts the collectively most useful subset of the clauses in several rounds (\emph{layers}) of
message passing. This means that approximative inference rounds done
by the GNN are efficiently interleaved with precise symbolic inference
rounds done inside E.
The GBDT and GNN versions have so far been used separately and only with CPU-based evaluation.
In this work, we add efficiently implemented GPU-based evaluation for the GNN and start to use the two methods cooperatively.

\section{Cooperative Filtering: Faster and Smarter}
\label{sec:filtering}
The set of generated clauses in saturation-style ATPs typically grows 
quadratically with the number of processed clauses. Each new given
clause is combined with all compatible previously processed clauses,
followed by (possibly expensive) evaluation of all newly generated
clauses. In particular, the GNN predictors typically incur a significant evaluation cost per clause.
The quadratic growth means that longer ENIGMA-GNN runs may get very slow.

To avoid large memory consumption and similar expensive evaluations in
long hint-based Prover9 runs (often taking several days) on the AIM
problems~\cite{KinyonVV13}, Veroff has used weight-based filtering,
discarding immediately clauses that reach a certain weight limit. This
often helps, but counterexamples are common, and in practice, such
schemes often need to be made more complicated.\footnote{We thank Bob
  Veroff for explaining that this is done by gradually lowering the
  weight limit inside a single longer Prover9 run, and by raising the
  initial weight limit and slowing down the weight reduction scheme
  across multiple Prover9 runs.} The three methods that we introduce
below are instead targeting this issue by using faster learning-based
filtering.

\subsection{Fast GNN Evaluation Using a GPU Server}
\label{server}

The main weakness of the GNN version of ENIGMA is its slow clause evaluation. %
In our previous ENIGMA Anonymous
experiments\cite{DBLP:conf/cade/JakubuvCOP0U20}, we used GPUs for model
training, but during the proof search we evaluated the clauses on a single CPU
(per each E prover's instance).
This was partly to provide a fair comparison with GBDTs which we also evaluate on
a single CPU, but also to avoid large start-up overheads when loading the neural models to
a GPU and running with low time limits.
Here we instead develop a persistent multi-threaded GPU server that evaluates clauses from multiple E prover runs using multiple GPUs.

The modification is as follows.
During the proof search, after computing the tensor representation of the
newly generated clauses, an E Prover client sends the tensors (in a JSON text
format) over a network socket to a remote server. The client then waits for the server response
which provides the scores (GNN evaluations) of the new clauses. This means that the clients are inactive for some time and more of them are needed to saturate the CPUs on the machines (see the detailed experimental discussion in Section~\ref{exp:server}).
This is typically not a problem due to many instances of E running with different premises and parameters in hammering and CASC LTB scenarios, as well as in many iterations of the learning/proving loop that attempt to solve harder and harder problems over a large problem set.

The remote server, written in Python, is launched before the E clients,
loading the GNN model to the (multiple) GPUs in advance. %
Once the model is loaded to the GPUs, the server accepts tensor queries on a designated
port, evaluates them on the GPUs, and sends the clause evaluations back to the
clients.
In more detail, the server is parameterized by the number $N$ (our
default is 28) of independent worker threads, the batch size $b$ (our
default is 8) and the waiting time $T$ (our default is 0.01s).  The
client queries are accumulated in a shared queue that the $N$ worker
threads process. Each worker operates in two steps. First, it checks
the queue, and if it contains less than $b$ queries, it waits for $T$
seconds. Then it evaluates the first $b$ queries on the queue, or less
if there are not enough of them available. Note that when the worker
waits or evaluates queries, other workers can process the queue.

The advantage is that the single GNN server amortizes the startup costs and handles queries of many E prover
clients and distributes them across multiple GPUs. This means that much larger batches (containing clauses coming from multiple clients)
are typically loaded onto the GPUs, amortizing also the relatively high cost of communication with the GPUs.
This results in large real time speed-ups over the CPU version, see
Section~\ref{exp:server}.
In our experiments, we run the GPU server and the E clients on the same machine.
Hence the network overhead is low because the communication is done over a local loopback interface.
In the case of a remote connection, the architecture would benefit from data
compression and/or binary data formats to decrease the network overhead.
See Section~\ref{exp:server} for the current average sizes of the data exchanged.

\subsection{Best of Both Worlds: GNN with GBDT Filtering}
\label{gnn-and-gbdt}

While the GPU server evaluation provides a considerable speed up, the
evaluation of clauses on a GPU is still relatively costly compared to the GBDT clause evaluation.
Hence we develop the following combination of the two methods, where the GBDT is used to pre-filter the clauses for the GNN.

In more detail, the set of clauses to be evaluated by the GNN is first evaluated by a fast GBDT model.\footnote{This feature  is implemented for the LightGBM models, which seem more easily tunable for such tasks.}
The GBDT model assigns a score between $0$ and $1$ to each clause, and only the
clauses with scores higher than a selected threshold are sent to the GPU server
for evaluation by the GNN.
The clauses which are filtered out by the GBDT model are assigned a very high
weight inside E Prover, which makes them unlikely to ever be selected for
processing. This way we prevent E from incorrectly reporting satisfiability when the good clauses run out.

Several requirements must be met for this filtering to be effective.
First, the GBDT filtering model must be small enough so that the evaluation is
fast, yet precise enough so that the more important clauses are not mistakenly filtered out too often.
Second, the score threshold must be properly fine-tuned, which typically requires 
experimental grid search on smaller samples. %
Experiments with a GBDT pre-filtering for a GNN are presented in
Section~\ref{exp:2phase}.

\subsection{Parental Guidance: Pruning the Given Clause Loop}
\label{parental}

We define \emph{(clausal) parental guidance} as clause evaluation based on the
features of the parents of a clause rather than on the clause itself.
Such fast rejection filters often help: in nature, mating is typically
highly restricted by various features of parents (e.g., their age, appearance,
finances, etc.). Similarly,  it does not often happen that clauses from very
different parts of mathematics (e.g., differential geometry and graph
theory) need to be resolved.

Parental guidance can be seen as ``just another filter'' of the
generated clauses, but its motivation is more radical: The ``good old''\footnote{The given clause loop is almost 50 years old as of 2021.}
given clause loop~\cite{Overbeek:1974:NCA:321812.321814} insists, for completeness reasons, on performing all
possible inferences between the processed clauses and the given
clause, typically leading to a quadratic growth of the set of
generated clauses. However, if we had perfect information about the
proof, this would be wasteful and could be replaced by just 
performing the inferences needed for the proof in each given clause
loop. With parental guidance, we instead propose to prune the given clause loop in a soft way: a trained predictor
judges the likelihood of the particular inference being needed for the
proof. When an inference is deemed useless, %
the clause is still
generated but immediately \emph{frozen} so that it does not have to
be evaluated by additional heuristics. %

The parental guidance is implemented using GBDTs (our \emph{parental model}), and the filter is directly put inside
E's given clause loop as follows.
When E selects a given clause $g$, E uses term indexes to efficiently determine which clauses can be combined with $g$ to %
generate new clauses. %
After generating the clauses, E performs simplifications, removes trivial clauses, evaluates the remaining clauses with the 
clause evaluation functions, and inserts them into the unprocessed set.  
The call to the parental model is
executed after the clause generation and prior to the simplifications.  
Clauses generated by paramodulation, which also implements resolution in E, have 
two parents, and these are judged by the parental model.  
Clauses whose parents are jointly scored below a chosen threshold are put into 
the \emph{freezer} set to avoid impairing the completeness of the proof search.
Clauses with good parents continue on to the unprocessed set.  
In case the unprocessed set becomes empty, the frozen clauses are revived and treated as usual.

Note that a naive alternative way to implement parental guidance would be to evaluate each given clause's compatibility 
with all previously processed clauses.  This would, however, result in many unnecessary GBDT queries and evaluations.
Instead, our approach allows E's indexing to find the typically much smaller
set of potential inferences and to limit the parental evaluation to them.\footnote{The
  efficiency boost obtained by using intelligent indexing
  is analogous to the boost obtained by using our structure-aware GNN
  for context-based neural clause selection (Section~\ref{enigma})
  rather than off-the-shelf Transformer models. The latter would
  quadratically consider interactions of all symbols in the context
  and query clauses, decreasing the evaluation speed by orders of
  magnitude, resulting in a very inefficient prover.}

There are various ways to represent the pair of parent clauses for the learning of the parental model.
In this work, we evaluate two methods: 
\begin{enumerate}
  \item $\ModelParM$ merges the feature vectors of the parent clauses into one vector, typically by simply adding
the feature counts\footnote{In some special cases of features, we instead take their maximum/minimum.}
  \item $\ModelParC$ concatenates the feature vectors of the parent clauses to preserve their information in full.
\end{enumerate}
An interesting future alternative is to include the difference of 
the parents' feature vectors in addition to their union and concatenation, 
which allows the GBDT to choose the most informative features.

\section{Experimental Setting and Baselines}
\label{setting}

\subsection{Evaluation Problems and Training Data}

All our experiments are performed%
\footnote{On a server with 36 hyperthreading Intel(R)
Xeon(R) Gold 6140 CPU @ 2.30GHz cores, 755 GB of memory, and 4 NVIDIA GeForce GTX 1080 Ti GPUs.} 
on a large benchmark of $\num{57880}$ problems%
\footnote{\url{http://grid01.ciirc.cvut.cz/~mptp/1147/MPTP2/problems_small_consist.tar.gz}}
originating from the Mizar Mathematical Library (MML)~\cite{KaliszykU13b}
exported to first-order logic by
MPTP~\cite{Urban06}.
We make use of our ongoing extensive evaluation of many AI/TP methods over this corpus%
\footnote{\url{https://github.com/ai4reason/ATP_Proofs}} that measures the overall 
improvement on this large dataset over the last similar evaluation done in~\cite{KaliszykU13b}. %
In these experiments we have significantly extended our previously published
results~\cite{DBLP:conf/cade/JakubuvCOP0U20}.\footnote{The publication of this large evaluation is in preparation.}
Proofs of 73.5\% (more than \num{40}k)
Mizar problems have been so far found by learning-guided ATPs, and numerous GBDT and GNN models for ATP
guidance have been trained.

In that experiment, all Mizar problems\footnote{\url{http://grid01.ciirc.cvut.cz/~mptp/Mizar_eval_final_split}}
are split (in a 90-5-5\% ratio) into \num{3} subsets: 
(1) \num{52}k problems for \emph{training},
(2) \num{2896} problems for \emph{development}, and 
(3) \num{2896} problems for final evaluation (\emph{holdout}).
We use this split here, and additionally we use a random subset of \num{5792} of the training problems to
speed up the training of various experimental methods.

\subsection{Baseline ENIGMA Models}

Out of the \num{52}k training problems, we were previously able to
prove more than \num{36}k problems, obtaining varied numbers of proofs
for each problem (ranging from $1$ to hundreds). On these \num{36}k
problems we train our baseline GBDT and GNN predictors.  To balance
the contribution of different problems during the training of the
predictors, we randomly choose at most 3 proofs for every proved
training problem. This yields a set of about \num{100}k
proofs, denoted further as the \emph{large} (training) set.
When limited to the \num{5792} random subset of the training problems, this yields \num{11748} proofs, denoted further as the \emph{small} training set.

On the \emph{large} set  we train the first baseline predictor denoted by
$\ModelLargeLgb$. This is a GBDT model (implemented by the LightGBM
framework) trained using the ENIGMA Anonymous clause representation
(Section~\ref{enigma}). %
The model consists of \num{150} decision trees of depth \num{40} with
\num{2048} leaves.  This model was selected as it performed best in
our previous experiments with standard GBDTs, being able to prove \num{1377} of the
\emph{holdout} problems using a 5 second limit per problem.  
Additionally, we train another model
$\ModelSmallLgb$ only on the \emph{small} set of training problems.
The model $\ModelSmallLgb$ is a LightGBM model with
\num{150} trees of depth \num{30} and with \num{9728} leaves.  The
training of $\ModelLargeLgb$ took around \num{27} minutes and the
training of $\ModelSmallLgb$ around \num{10} minutes, both on \num{30}
CPUs. These are relatively low and practical times compared to the
training of neural networks.

We also train baseline GNN models on the same data, denoted
$\ModelLargeGnn$ and $\ModelSmallGnn$ respectively.  The training of
$\ModelLargeGnn$ for 45 epochs takes about 15 hours on the full set of
100k proofs on a high-end NVIDIA V100 GPU card.\footnote{We use the same GNN hyper-parameters as in~\cite{DBLP:conf/ecai/OlsakKU20,DBLP:conf/cade/JakubuvCOP0U20} with the exception of the number of \emph{layers} that we increase here to \num{10}.}
It would likely
take days when training with CPUs only.  We choose for the ATP
evaluation the (39th) snapshot that achieves both the best loss
(0.2063) and the best weighted accuracy (0.9147) on 5\% of the data
that we do not use for training.
The training of $\ModelSmallGnn$ for 100
epochs takes about 4 hours on the \emph{small} set using the same GPU card.  We
choose for the ATP evaluation the (56th) snapshot that
achieves the best loss (0.2988) on 5\% of the data that we do not
use for training.  The weighted accuracy on this set is 0.8685, which
is also among the highest values.

In the evaluation we run all our baseline ENIGMA predictors in an
equal combination with a strong non-learning E strategy $\S$ (see Appendix~\ref{sec:str}).
This means that the processed clauses are selected in (equal) turns by
ENIGMA and by $\S$. This \emph{coop} mode has typically worked better
than the \emph{solo} mode, where only the ENIGMA predictor is doing
the clause selection.

\subsection{Training of the Parental GBDT Models}
The training data for the parental guidance models are generated by running E
using either $\ModelLargeLgb$ or $\ModelLargeGnn$ on the \num{52}k \emph{training} problems with a 30
second time limit and by printing 
the derivation
of all clauses generated during the proof search.%
\footnote{Using E's option ``\texttt{--full-deriv}''.}
We considered the following two schemes to classify the good pairs of parents and to generate the training data:
\begin{enumerate}
  \item $\ModelParP$ classifies parents of only the proof clauses as \emph{positive} and all other generated clauses as \emph{negative}.
  \item $\ModelParS$ classifies parents of all processed (selected) clauses as \emph{positive} and the unprocessed generated clauses as \emph{negative}.
\end{enumerate}
The rationale behind $\ModelParP$ is that every non-proof clause should be pruned if possible.  
The rationale behind $\ModelParS$ is that if an effective clause selection strategy, such as $\ModelLargeLgb$, 
predicted a clause to be useful, then it is probably worth generating. However, such data may be confusing as it 
includes clauses that did not contribute to the proof.  

If a pair of parents produces both positive and negative clauses, we consider the pair positive in our implementation. 
However, this does not happen very often.
Based on a survey on the \emph{small}
set labeled according to $\ModelParMP$, 73\% of the problems have no conflict. %
There are \num{1519} parents of both positive and negative clauses, \num{53359} are positive, and \num{6086414} are negative. 
Under $\ModelParMS$, \num{9798} of the parents are mixed, \num{854778} are positive, and \num{5178592} are negative.  
In either case, the primary learning task is to identify and prune as many negative clauses as possible without filtering 
a necessary proof clause by mistake.

One parameter to experimentally tune is the \emph{pos-neg ratio} used in the
GBDT training: the ratio of positive and negative examples.
The pos-neg ratio is \num{1}:\num{192} over the \emph{large} $\ModelParMP$ data, which is more than
ten %
times more than the ratio of the training data for $\ModelLargeLgb$ and $\ModelLargeGnn$.
Hence, reducing the pos-neg ratio by randomly sampling negative examples could further
boost the training performance.

The parental guidance models are trained using GBDTs.
Trained models are evaluated in combination with the 
GBDT or GNN clause evaluation heuristic using either the $\ModelLargeLgb$ or $\ModelLargeGnn$ model, see
Section~\ref{exp:parental}.

\section{Evaluation of the New Methods}
\label{evaluation}

\subsection{Speedup by Using a GPU Server }
\label{exp:server}

First we measure the speedup obtained by evaluating the ENIGMA GNN
calls on a separate GPU server.  To avoid network latency and for a cleaner comparison, we run both
the clients (E/ENIGMA) and the GPU server on the same machine equipped
with four NVIDIA GeForce GTX 1080 GPU cards and 36 hyperthreading
CPU cores. We configure the server to use all four GPU cards. Its other
important parameters are the number of worker threads and the batch
size. We experimentally set them to 28 and 8, and we use $\ModelLargeGnn$ for all proof runs.

Comparison of the CPU-only and GPU-server versions is
complicated by the fact that the server-based GNN evaluations do not
count towards the CPU time taken by E, as reported by the operating
system. Still, a comparison using the CPU time is interesting and we include
it, using 30 and 60 second CPU limits for the CPU-only version, and a 30 second CPU limit for the client-server version.

Another way to compare the two is by using parallelization, i.e., running many
instances of E in parallel. In the client-server version
the instances talk to the GPU server simultaneously. We saturate the machine's
CPUs fully for both versions, and run for approximately
equal overall real time over the development and holdout sets. This is
roughly achieved by using 60s time limit with 70-fold parallelization
for the CPU version, and 30s time limit with 160-fold parallelization
for the client/server version. The CPU version then takes about \num{27.5}
minutes to finish on the \num{2896} problems, while the client-server takes
about 34 minutes to finish.  Table~\ref{cpu-gpu} compares the number
of solved problems on the development and holdout sets. The GPU server improves the performance on
the development resp. holdout sets by 9.5\% resp. 11.5\%.

We also
compare the average number of generated clauses on the problems that
timed out in both versions. In the 60s CPU version it is \num{16835}, while in
the 30s client-server it is \num{63305}. This is a considerable speedup,
achieved by employing the additional custom hardware---our
four GPU cards. The average number of GNN queries in the 1358 problems
that timed out in the 30s GPU server runs is 243.8, and on average the
communication with the GPU server took 155MB in a timed-out problem. A
single GNN query took on average 637kB.

\begin{table}[!htbp]
\begin{small}
\caption{\label{cpu-gpu} Comparison of the CPU-only GNN ENIGMA with the client-server version using GPUs. 
All runs are evaluating $\ModelLargeGnn$ on the whole development (D) and holdout (H) datasets. 
The percentage improvement is computed over the 60s CPU version that corresponds more closely in real time to the client-server version. 
All runs use queries of size 256 and contexts of size 768.}%
\begin{center}
  \begin{tabular}{lllllll}
    \toprule
    set & model & method & time &   solved\\
    \midrule                                           
D & \ModelLargeGnn & CPU & 30 &    1311       \\                                                     
D& \ModelLargeGnn & CPU & 60 &  1380         \\              
D & \ModelLargeGnn & GPU & 30 &     1511     (+9.5\%)   \\
   \bottomrule
  \end{tabular}
\quad
  \begin{tabular}{lllllll}
    \toprule
    set & model & method & time &   solved\\
    \midrule
H& \ModelLargeGnn& CPU & 30 & 1301         \\              
H& \ModelLargeGnn& CPU & 60 & 1371         \\              
H &\ModelLargeGnn& GPU & 30 &    1529     (+11.5\%)   \\
   \bottomrule
  \end{tabular}
\end{center}
\end{small}
\vspace{-6mm}
\end{table}

\subsection{Evaluation of 2-phase ENIGMA}
\label{exp:2phase}

\subsubsection{Small GBDT and Small GNN:}

In the first experiment we use the GBDT and GNN predictors $\ModelSmallLgb$ and $\ModelSmallGnn$ trained on the
\emph{small} subset of the training dataset.

We first do a grid search over the parameters on a smaller dataset of
300 development problems (see Table~\ref{Grid1ss} in Appendix~\ref{sec:app} for the full grid search).
Then we evaluate the
best parameters on the development and holdout sets and compare them with
the standalone performance of $\ModelSmallGnn$, which is the stronger of the two baselines (Table~\ref{Grid2ss}).
The best combined methods are then evaluated also in 60s. This gives a
relatively fair real-time comparison to the standalone GNN, because
the reported CPU times do not include the time taken by the GPU
server.\footnote{We have made this estimate based on a comparison of
  real and CPU times done on a set of problems that time out in both
  methods.}

  Our best combined method solves (in real time) 10.4\%,
  resp.\ 9.0\%, more problems on the development, resp.\ holdout, set than
the standalone GNN. This is a significant improvement, which will
likely get even more visible with higher time limits, because of the
quadratic growth of the set of generated clauses. The performance
improvement over the standalone GBDT model is even larger.
\begin{table}[!htbp]
  \vspace{-4mm}
\caption{\label{Grid2ss} Final evaluation of the best combination of $\ModelSmallLgb$ with $\ModelSmallGnn$ on the whole development (D) and holdout (H) datasets.}
\begin{center}
  \begin{small}
      \vspace{-1mm}
  \begin{tabular}{lllllll}
    \toprule
    set & model & thresh. & time &  query &   context & solved\\
    \midrule                                           
D & \ModelSmallGnn & -   & 30 & 256 & 768  &   1251       \\                                                     
D & \ModelSmallLgb & -   & 30 & - & -  & 1011         \\              
D & \ModelSmallLgb+\ModelSmallGnn\ \ \ & 0.01 & 60 & 512 & 1024  &    1381     (+10.4\%)   \\
D & \ModelSmallLgb+\ModelSmallGnn & 0.03 & 60 & 512 & 1024  &    1371     (+9.6\%)   \\
D & \ModelSmallLgb+\ModelSmallGnn & 0.03 & 30 & 512 & 1024  &    1341     (+7.2\%)   \\
D & \ModelSmallLgb+\ModelSmallGnn & 0.01 & 30 & 512 & 1024  &    1339     (+7.0\%)   \\
    \midrule
H& \ModelSmallGnn &  -   & 30 & 256 & 768  & 1277         \\
H& \ModelSmallLgb &  -   & 30 & - & -  & 1002         \\              
H& \ModelSmallLgb+\ModelSmallGnn& 0.01 & 60 & 512 & 1024  &    1392   (+9.0\%)   \\
H& \ModelSmallLgb+\ModelSmallGnn& 0.03 & 60 & 512 & 1024  &    1387   (+8.6\%)   \\
H& \ModelSmallLgb+\ModelSmallGnn& 0.01 & 30 & 512 & 1024  &    1361   (+6.6\%)   \\
H& \ModelSmallLgb+\ModelSmallGnn& 0.03 & 30 & 512 & 1024  &    1353   (+6.0\%)   \\
   \bottomrule
  \end{tabular}
\end{small}
\end{center}
\vspace{-5mm}
\end{table}

\subsubsection{Large GBDT and Small GNN:}
In the next experiment, we want to see how much the training of the less expensive
model (GBDT) on more data helps. I.e., we replace $\ModelSmallLgb$ with $\ModelLargeLgb$ and keep $\ModelSmallGnn$.
This has practical applications in
real time, because cheaper ML predictors such as GBDTs are faster to
train than more expensive ones such as the GNN.
We again first do a grid search over the parameters on a small dataset
of 300 development problems (see Table~\ref{Grid1ls} in Appendix~\ref{sec:app}).
Then we evaluate
the best models on the development and holdout sets and compare them
with the standalone performance of $\ModelLargeLgb$ and $\ModelSmallGnn$
(Table~\ref{Grid2ls}).  The best combined methods are then again
evaluated also in 60s, which makes it comparable in real time to the
standalone GNN model.

\begin{table}[!htbp]
\caption{\label{Grid2ls} Final evaluation of the best combination of $\ModelLargeLgb$ and $\ModelSmallGnn$  on the whole development (D) and holdout (H) datasets.}
\begin{small}
\begin{center}
  \begin{tabular}{lllllll}
    \toprule
    set & model & thresh. & time &  query &   context & solved\\
    \midrule                                    
    D     & \ModelSmallGnn & - & 30 & 256 & 768 & 1251 \\
    D     & \ModelLargeLgb & - & 30 & - & -  & 1397         \\
    D     & \ModelLargeLgb+\ModelSmallGnn\ \ \ & 0.3 & 60 & 2048 & 768 & 1527 (+9.3\%) \\
    D     & \ModelLargeLgb+\ModelSmallGnn & 0.3 & 30 & 2048 & 768 & 1496 (+7.1\%)\\        
    \midrule
      H  & \ModelSmallGnn & - & 30 & 256 & 768 & 1277 \\
      H  & \ModelLargeLgb & - & 30 & - & -  & 1390         \\
      H  & \ModelLargeLgb+\ModelSmallGnn & 0.3 & 60 & 2048& 768 & 1494 (+7.5\%)\\
      H  & \ModelLargeLgb+\ModelSmallGnn & 0.3 & 30 & 2048& 768 & 1467 (+5.5\%)\\      
   \bottomrule
  \end{tabular}
\end{center}
\end{small}
\vspace{-6mm}
\end{table}

Our best combined method solves (in CPU time)
7.1\%, resp.\ 5.5\%, more problems on the development, resp.\ holdout, set
than the standalone GBDT. For the GNN, this is (in real time) 9.3\%
resp.\ 7.5\%.  These are smaller gains than in the previous $\ModelSmallLgb+\ModelSmallGnn$ %
scenario, most likely because the stronger predictor dominates here.
Also note that the large query ($2048$) used in our strongest model is
typically diminished a lot by the GBDT pre-filter, resulting in average
query sizes after the GBDT pre-filtering of 256--512.

\subsubsection{Large GBDT and Large GNN:}
Finally, we evaluate the large setting, using
the GBDT and GNN predictors $\ModelLargeLgb$ and $\ModelLargeGnn$ trained on the
full training dataset.
Again, we first do a grid search over the parameters on the small
set of 300 development problems (Table~\ref{Grid1ll} in Appendix~\ref{sec:app}).
Then we evaluate
the best parameters on the development and holdout sets, and we compare them
with the standalone performance of $\ModelLargeLgb$ and $\ModelLargeGnn$ 
(Table~\ref{Grid2ll}). The improvements on the development, resp.\ holdout, set
is 9.1\%, resp.\ 7.3\%, in real time, and 6.9\%, resp.\ 4.8\%, when using CPU time.
The E auto-schedule solves in 30s (CPU time) 1020 of the holdout problems. Our strongest 2-phase method solves 1602 of these problems in the same CPU time, i.e., 57.1\% more problems.
\begin{table}[!htbp]
\caption{\label{Grid2ll} Final evaluation of the best combination of $\ModelLargeLgb$ and $\ModelLargeGnn$ on the whole development (D) and holdout (H) datasets.}
\begin{small}
\begin{center}
  \begin{tabular}{lllllll}
    \toprule
    set & model & thresh. & time &  query &   context & solved\\
    \midrule                                    
    D & \ModelLargeGnn & -   & 30 & 256 & 768 & 1511 \\
    D & \ModelLargeLgb & -   & 30 & - & -  & 1397         \\
    D & \ModelLargeLgb+\ModelLargeGnn\ \ \ & 0.1 & 60 & 1024 & 768 & 1648 (+9.1\%) \\
    D & \ModelLargeLgb+\ModelLargeGnn & 0.1 & 30 & 1024 & 768 & 1615 (+6.9\%)\\    
    \midrule
    H & \ModelLargeGnn & -   & 30 & 256 & 768 & 1529 \\
    H & \ModelLargeLgb & -   & 30 & - & -  & 1390         \\
    H & \ModelLargeLgb+\ModelLargeGnn & 0.1 & 60 & 1024& 768 & 1640 (+7.3\%)\\
    H & \ModelLargeLgb+\ModelLargeGnn & 0.1 & 30 & 1024& 768 & 1602 (+4.8\%)\\
   \bottomrule
  \end{tabular}
\end{center}
\end{small}
\vspace{-6mm}
\end{table}

\subsection{Evaluation of the Parental Guidance Combined with  $\ModelLargeLgb$}
\label{exp:parental}
The parameters for parental guidance models are explored via a series of grid searches to reduce the number of combinations.
Initially, we only use $\ModelLargeLgb$ in conjunction with the parental models.
First, the training data classification schemes, $\ModelParMP$ and $\ModelParMS$, are compared with a grid search over the pos-neg reduction ratio.    
The best combination of reduction ratio and classification scheme is used to perform a grid search over LightGBM parameters for $\ModelParM$.    
Next, reduction ratio and LightGBM parameter grid searches are done with the $\ModelParC$ featurization method data, 
starting with the best $\ModelParM$ parameters from the previous experiments.  
Every model is evaluated with the same set of nine parental filtering thresholds $\{0.005, 0.01, 0.03, 0.05, 0.1, 0.2, 0.3, 0.4, 0.5\}$.  
The grid searches are done over the \num{300} problem development set and run for \num{30} seconds.  
On this dataset, $\ModelLargeLgb$ solves 159 problems.  

\subsubsection{Pos-neg reduction ratio tuning (merge):}
The first grid search examines the pos-neg reduction ratio denoted as $\rho$.
Before the reduction, the average pos-neg ratio for $\ModelParMS$ is $1:9.2$ and the average for $\ModelParMP$ is $1:191.8$.
We reduce the pos-neg ratio to a given $\rho$ by randomly sampling the negative examples on a problem-specific basis. 
This means that the average pos-neg ratio over the whole dataset is typically a bit smaller than $\rho$.
For example, using $\rho=4$ on the $\ModelParMP$ results in 
an average of \num{3.95} times more negative than positive examples.  
Both $\ModelParMS$ and $\ModelParMP$ are tested using $\rho\in\{-, 1, 2, 4, 8, 16\}$
 where ``$-$'' denotes using 
the full training dataset.
We use the best LightGBM model parameters discovered during prototyping of the parental guidance features: 
the parameters are \num{50} trees of depth \num{13} with \num{1024} leaves. %
\begin{table}[!htbp]
\centering
  \caption{\label{Gridrr} The best threshold for each tested reduction ratio.
            The threshold of 0.03 was identical to 0.05 for all tested ratios with $\ModelParMS$, 
            whereas there are no ties among thresholds for $\ModelParMP$.}
\centering
  \begin{tabular}{l|clllll}
    \toprule
\ensuremath{\rho_\textsf{fuse}^\textsf{given}}
    & $-$        & 1     & 2     & 4    & 8    & 16 \\
    \midrule
    threshold & 0.05 & 0.05  & 0.05  & 0.05 & 0.05 & 0.05 \\
    solved    & 161  & 161   & 161   & 161  & 161  & 160 \\
       \bottomrule
    \end{tabular}
    \quad
  \begin{tabular}{l|clllll}
    \toprule
    \ensuremath{\rho_\textsf{fuse}^\textsf{proof}}
    & $-$        & 1     & 2     & 4    & 8    & 16 \\
    \midrule
    threshold & 0.005 & 0.2  & 0.2  & 0.2 & 0.2 & 0.2 \\
    solved    & 111  & 164   & 163   & 165  & 162  & 164 \\
       \bottomrule
    \end{tabular}
\end{table}

Table~\ref{Gridrr} shows that the reduction ratio makes significant difference for the $\ModelParMP$ data 
and almost none for $\ModelParMS$ data, which is probably because the $\ModelParMS$ data are already reasonably balanced.
Moreover, parental guidance seems to perform better with $\ModelParMP$ data than $\ModelParMS$ data, %
probably because mistakes of $\ModelLargeLgb$ are included in the training data.  
In the following experiments, only the $\ModelParP$ classification scheme is used (so the prefix is dropped).

\subsubsection{LightGBM parameter tuning (merge):}
Next we perform the second grid search over the LightGBM training hyper-parameters for $\ModelParM$,
fixing $\rho=4$ as it performed best.  
We try the following values for the three main hyper-parameters, namely, for the
number of trees in a model, the maximum number of tree leaves, and the maximum
tree depth:
\begin{eqnarray*}
   \mathrm{trees}  & \in & \{ 50   , 100  , 150               \} \\
   \mathrm{leaves} & \in & \{ 1024 , 2048 , 4096 , 8192 , 16384 \} \\
   \mathrm{depth}  & \in & \{ 13   , 40   , 60   , 256         \} 
\end{eqnarray*}
The best model for $\ModelParM$ solves \num{171} problems and consists of
\num{100} trees, with the depth \num{40}, and \num{8192} leaves, and a threshold
of \num{0.05}.  
Another eight models solve \num{169} problems.  
We also tested these parameters to find a better model for $\ModelParMS$, %
which solves \num{163} problems with $\rho=8$ and a threshold of \num{0.1}.

\begin{table}[tbp]
\centering
  \caption{\label{Gridrrc} The best threshold for each tested reduction ratio of $\ModelParC$.  
                           }
  \begin{tabular}{l|clllll}
    \toprule
\ensuremath{\rho_\textsf{cat}}
    & $-$ & 1   & 2    & 4   & 8   & 16 \\
    \midrule
    threshold           & 0.5 & 0.1 & 0.05 & 0.3 & 0.1 & 0.05 \\
    solved              & 117 & 168 & 170  & 168 & 173 & 169 \\
       \bottomrule
    \end{tabular}
\end{table}

\subsubsection{Pos-neg reduction ratio tuning (concat):}
This grid search uses the best LightGBM hyper-parameters for $\ModelParM$ to test the same reduction ratios 
and thresholds for $\ModelParC$. 
Table~\ref{Gridrrc} shows that $\ModelParC$ outperforms $\ModelParM$ 
and $\rho=8$ is the best. Reducing the negatives is even more important here.

\subsubsection{LightGBM parameter tuning (concat):}
\begin{figure}[!h]
    \begin{centering}
    \includegraphics[scale=0.50]{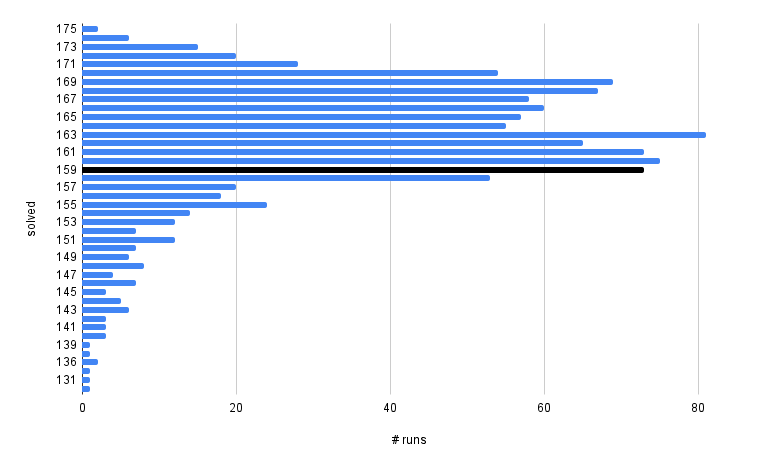}
   \caption{\label{gridcountc} The number of settings (and runs) corresponding to each number of solutions for the $\ModelParC$ grid search. 
            The black bar is \num{159}, the number of problems solved by $\ModelLargeLgb$.
            Only \num{154} (20\%) of the runs interfere with $\ModelLargeLgb$'s performance and solve fewer problems.  
            These runs largely consist of the thresholds, $\{0.3, 0.4, 0.5\}$, 
            but the only parameter whose majority of runs score below $\ModelLargeLgb$ is a threshold of \num{0.5}.  
            The outliers tend to be larger models. }
    \end{centering}
\end{figure}

The grid search for the $\ModelParC$ data is done over the following hyper-parameters:
\begin{eqnarray*}
   \mathrm{trees}  & \in & \{ 50   , 100  , 150, 200               \} \\
   \mathrm{leaves} & \in & \{ 1024 , 2048 , 4096 , 8192 , 16384, 32768 \} \\
   \mathrm{depth}  & \in & \{ 13   , 40   , 60   , 256, 512         \} 
\end{eqnarray*}
The upper limits have increased compared to the $\ModelParM$ grid-search 
because one of the best models had \num{150} trees of depth \num{256}, placing it at the edge of the grid.  
The best models solve \num{174}-\num{175} problems.  
These are evaluated on the full development set (Table~\ref{Gridclgb}).  
The larger models seem to work best with a threshold of \num{0.05} and the smaller models with a threshold of \num{0.2}, which is likely because they can be less precise.
The full distribution of the results can be seen in Figure~\ref{gridcountc}.  
The number of parameter configurations that outperform the baseline suggests that parental guidance is an effective method.

\begin{table}[htbp]
\centering
\caption{\label{Gridclgb} The best $\ModelParC$ models with $\rho=8$.}
\centering
  \begin{tabular}{llllll}
    \toprule
    trees & depth & leaves   & threshold\ \ \ & solved (300) & solved (D) \\
    \midrule                                                                                   
    200 & 60  & 4096 & 0.05 & 175 & 1557  \\
    200 & 512 & 4096 & 0.05 & 175 & 1561 \\
    200 & 256 & 4096 & 0.05 & 174 & 1558 \\
    150 & 512 & 1024 & 0.2  & 174 & 1568 \\
    150 & 256 & 1024 & 0.2  & 174 & 1556 \\
    100 & 60  & 8192 & 0.05 & 174 & \textbf{1571} \\
    100 & 40  & 2048 & 0.2  & 174 & 1544 \\
    100 & 40  & 2048 & 0.1  & 174 & 1544 \\
   \bottomrule
  \end{tabular}
\vspace{-8mm}
\end{table}

\begin{table}[htbp]
\centering
\caption{\label{Gridpgfinal}Final 30s evaluation on small trains (T), 
         development (D), and holdout (H) compared with $\ModelLargeLgb$.}
\centering

  \begin{tabular}{lllll}
    \toprule
    model   & threshold\ \ \ & solved (T) & solved (D) & solved (H) \\
    \midrule                                                                                   
   \ModelLargeLgb                     & - & 3269 & 1397 & 1390 \\     
   \ModelParMS+\ModelLargeLgb\ \ \ & 0.05 & 3302 (+1.0\%)\ \ \ & 1411 (+1.0\%)\ \ \ & 1417 (+1.9\%)\\
   \ModelParMP+\ModelLargeLgb      & 0.1  & 3389 (+3.7\%) & 1489 (+6.6\%) & 1486 (+6.9\%)\\
   \ModelParC+\ModelLargeLgb      & 0.05 & 3452 (+5.6\%) & 1571 (+12.4\%) & 1553 (+11.7\%)\\
   \bottomrule
  \end{tabular}
\vspace{-2mm}
\end{table}

Finally we evaluate the best models on the small training, development, and holdout sets, and we
compare them with the standalone performance of \ModelLargeLgb{}  (Table~\ref{Gridpgfinal}).
Parental guidance achieves a significant improvement in performance on all %
datasets, solving \num{11.7}\% more on the holdout set. %
It is interesting to note that the improvement is %
greater on the development and holdout sets than on the training set.
For parental guidance it seems superior to classify only \emph{proof clauses} as positive examples. %
This is most likely due to LightGBM being confused by processed clauses that did not contribute to any proof.  
The method of concatenating the parent clause feature vectors ($\ModelParC$) seems far superior to merging them ($\ModelParM$).  
This is likely because merging the features is lossy and the order of the parents matters when performing inferences.

The results indicate that pruning clauses prior to clause evaluation is helpful.  
ENIGMA models tend to run best in equal combination with a strong E strategy, but this means they have 
no control over 50\% of the clauses selected for processing.  The ability to filter which clauses 
the strong E strategy can evaluate and select may be part of the strength behind parental guidance.  

\subsection{Parental Guidance with  $\ModelLargeGnn$ and 3-phase ENIGMAs}
\label{exp:parentalgnn}

We also explore a limited number of the most useful
hyper-parameters from Sections~\ref{exp:parental} and \ref{exp:2phase}
to combine the parental filtering with ENIGMA-GNN using
$\ModelLargeGnn$ and to create a 3-phase ENIGMA. We train a new
LightGBM parental filtering model on the $\ModelParC$ data generated by
running $\ModelLargeGnn$, using $\rho=8$, $\mathrm{trees}=100$,
$\mathrm{leaves}=8192$, and $\mathrm{depth}=60$. The grid search on
the 300 development problems leads to the best threshold values of
$0.005$ and $0.01$ when using $\mathrm{context}=768$ and $\mathrm{query}=256$ for ENIGMA-GNN with $\ModelLargeGnn$.

The version with the $0.01$ threshold then reaches so far the highest
value of 1621 development problems in 30s CPU time. This is 50 more
than the best parental result using $\ModelLargeLgb$ and 6 more than
the best 2-phase result. On the holdout set this setting yields 1623
problems, i.e., 70 more than the best $\ModelLargeLgb$ parental result
and 21 more than the best 2-phase result.

Finally, we explore 3-phase ENIGMAs, i.e., combinations of all the
methods developed in this work. This means that we first use the
parental guidance filtering, followed by the 2-phase evaluation which
in turn uses the GPU server. This implies a higher evaluation cost,
since both the parental and the first-stage LightGBM models are loaded on startup and are used to filter the
clauses.

We only tune the parental threshold and context and query
values, keeping the 2-phase threshold fixed at $0.1$. The best result
is again obtained by setting the parental threshold to $0.01$,
$\mathrm{context}=768$ and $\mathrm{query}=256$. This solves 1631
resp. \textbf{1632} of the development resp. holdout problems in 30s
CPU time.  This is our ultimate result, which is exactly 60\%
higher than the 1020 problems solved by E's auto-schedule in 30s
CPU time. It is also 17.4\% higher than the best ENIGMA result
prior to this work (1390 by standalone $\ModelLargeLgb$).

\section{Conclusion and Examples}
We have described several additions to the ENIGMA system. The new
methods combine fast(er) and smart(er) clause evaluation using
ENIGMA's parameterizable learning-based setting.  The GPU server allows much faster
runs of the neurally-guided ENIGMA, improving its real-time performance by about 10\%.
The parental guidance allows one to train clause evaluation differently from standard ENIGMA, providing 
an improvement of \num{11.7}\% on the holdout set.
Both when training
on small and on large datasets, the 2-phase methods provide good
improvements on the holdout sets (9\% and 7.3\%) over the strongest
standalone methods. The methods are adjustable and they will likely
lead to even higher improvements in longer runtimes, due to the
typically quadratic growth of the set of generated clauses in
saturation-style ATPs.  Our strongest 3-phase method improves E's
auto-schedule on the holdout set by 60\% in 30 seconds and
our best prior ENIGMA result by 17.4\%.

Several examples of the new proofs produced only by the methods
developed here are available on our project's web
page. %
Theorem
\texttt{INTEGR13:27}\footnote{\url{https://github.com/ai4reason/ATP_Proofs/\#differentiation---cot--ln-x--1--x--sin-ln-x2-}}
about the differentiation of $-cot(ln(x))$ needed 3904 nontrivial
given clause loops and 38826 nontrivial generated clauses, taking only
18s with the 2-phase ENIGMA. This can be compared to the previous
related theorem
\texttt{FDIFF\_7:36}\footnote{\url{https://github.com/ai4reason/ATP_Proofs/\#differentiation-exp_r--cos--x----exp_r--cos--x--sin-x}}
(differentiation of $exp(cos(x))$) done in the old setting, taking
28.4s to do only 1284 nontrivial given clause loops and 13287
nontrivial generated clauses. Other examples include a 486-long
proof\footnote{\url{https://github.com/ai4reason/ATP_Proofs/\#integral-chi-aa-is-integrable--integral-chi-aa--vol-a-486-long-atp-proof-from-63-premises}}
of a theorem about integrals done only in 41s with the 2-phase ENIGMA
evaluating 100k clauses, or a 259-long computational
proof\footnote{\url{https://github.com/ai4reason/ATP_Proofs/\#17-is-prime}}
about Fermat primes found in 11s while evaluating 52k clauses. Such
proofs are found despite hundreds of redundant axioms, by using new combinations of faster and smarter
trained ENIGMAs that efficiently guide the search.

\section{Acknowledgments}

This work was partially supported by the ERC Consolidator grant \emph{AI4REASON} no.~649043 (ZG, JJ, and JU),
the
European Regional Development Fund under the Czech project AI\&Reasoning no. CZ.02.1.01/0.0/0.0/15\_003/0000466 (ZG, JU, KC), the ERC Starting
Grant \emph{SMART} no.~714034 (JJ, MO), and by the Czech MEYS under the ERC CZ project \emph{POSTMAN}
no.~LL1902 (JJ).

\begin{comment}
\end{comment}
\bibliographystyle{plain}
\bibliography{ate11,stsbib}

\begin{thebibliography}{10}

\bibitem{AAB:2016tensorflow-2016}
Mart{\i}n Abadi, Ashish Agarwal, Paul Barham, Eugene Brevdo, Chen Zhifeng,
  Craig Citro, Greg~S. Corrado, Andy Davis, Jeffrey Dean, Matthieu Devin~Sanjay
  Ghemawat, Andrew~Harp Ian~Goodfellow, Geoffrey Irving, Michael Isard,
  Yangqing Jia, Rafal Jozefowicz, Manjunath~Kudlur Lukasz~Kaiser, Josh
  Levenberg, Dan Mane, Rajat Monga, Sherry Moore, Derek Murray, Chris Olah,
  Mike Schuster, Jonathon Shlens, Benoit Steiner, Ilya Sutskever, Paul~Tucker
  Kunal~Talwar, Vincent Vanhoucke, Vijay Vasudevan, Fernanda Viegas, Oriol
  Vinyals, Pete Warden, Martin Wattenberg, Martin Wicke, Yuan Yu, , and
  Xiaoqiang Zheng.
\newblock {Tensorflow: Large-scale machine learning on heterogeneous
  distributed systems}.
\newblock {\em arXiv preprint arXiv:1603.04467}, 2016.

\bibitem{Chen:2016:XST:2939672.2939785}
Tianqi Chen and Carlos Guestrin.
\newblock {XGBoost}: A scalable tree boosting system.
\newblock In {\em Proceedings of the 22nd ACM SIGKDD International Conference
  on Knowledge Discovery and Data Mining}, KDD '16, pages 785--794, New York,
  NY, USA, 2016. ACM.

\bibitem{DBLP:conf/cade/ChvalovskyJ0U19}
Karel Chvalovsk{\'{y}}, Jan Jakub\r{u}v, Martin Suda, and Josef Urban.
\newblock {ENIGMA-NG:} efficient neural and gradient-boosted inference guidance
  for {E}.
\newblock In Pascal Fontaine, editor, {\em Automated Deduction - {CADE} 27 -
  27th International Conference on Automated Deduction, Natal, Brazil, August
  27-30, 2019, Proceedings}, volume 11716 of {\em Lecture Notes in Computer
  Science}, pages 197--215. Springer, 2019.

\bibitem{Fan:2008:LLL:1390681.1442794}
Rong-En Fan, Kai-Wei Chang, Cho-Jui Hsieh, Xiang-Rui Wang, and Chih-Jen Lin.
\newblock Liblinear: A library for large linear classification.
\newblock {\em J. Mach. Learn. Res.}, 9:1871--1874, June 2008.

\bibitem{gittins1979bandit}
John~C Gittins.
\newblock Bandit processes and dynamic allocation indices.
\newblock {\em J. the Royal Statistical Society. Series B (Methodological)},
  pages 148--177, 1979.

\bibitem{DBLP:conf/itp/GoertzelJ0U18}
Zarathustra Goertzel, Jan Jakub\r{u}v, Stephan Schulz, and Josef Urban.
\newblock {ProofWatch}: Watchlist guidance for large theories in {E}.
\newblock In Jeremy Avigad and Assia Mahboubi, editors, {\em Interactive
  Theorem Proving - 9th International Conference, {ITP} 2018, Held as Part of
  the Federated Logic Conference, FloC 2018, Oxford, UK, July 9-12, 2018,
  Proceedings}, volume 10895 of {\em Lecture Notes in Computer Science}, pages
  270--288. Springer, 2018.

\bibitem{GoertzelJU19}
Zarathustra Goertzel, Jan Jakub\r{u}v, and Josef Urban.
\newblock Enigmawatch: Proofwatch meets {ENIGMA}.
\newblock In Serenella Cerrito and Andrei Popescu, editors, {\em Automated
  Reasoning with Analytic Tableaux and Related Methods - 28th International
  Conference, {TABLEAUX} 2019, London, UK, September 3-5, 2019, Proceedings},
  volume 11714 of {\em Lecture Notes in Computer Science}, pages 374--388.
  Springer, 2019.

\bibitem{DBLP:conf/gcai/2015}
Georg Gottlob, Geoff Sutcliffe, and Andrei Voronkov, editors.
\newblock {\em Global Conference on Artificial Intelligence, {GCAI} 2015,
  Tbilisi, Georgia, October 16-19, 2015}, volume~36 of {\em EPiC Series in
  Computing}. EasyChair, 2015.

\bibitem{Hillenbrand03}
Thomas Hillenbrand.
\newblock Citius altius fortius: Lessons learned from the theorem prover
  {WALDMEISTER}.
\newblock {\em ENTCS}, 86(1):9--21, 2003.

\bibitem{DBLP:conf/cade/JakubuvCOP0U20}
Jan Jakub\r{u}v, Karel Chvalovsk{\'{y}}, Miroslav Ols{\'{a}}k, Bartosz
  Piotrowski, Martin Suda, and Josef Urban.
\newblock {ENIGMA} anonymous: Symbol-independent inference guiding machine
  (system description).
\newblock In Nicolas Peltier and Viorica Sofronie{-}Stokkermans, editors, {\em
  Automated Reasoning - 10th International Joint Conference, {IJCAR} 2020,
  Paris, France, July 1-4, 2020, Proceedings, Part {II}}, volume 12167 of {\em
  Lecture Notes in Computer Science}, pages 448--463. Springer, 2020.

\bibitem{JakubuvU17a}
Jan Jakub\r{u}v and Josef Urban.
\newblock {ENIGMA:} efficient learning-based inference guiding machine.
\newblock In Herman Geuvers, Matthew England, Osman Hasan, Florian Rabe, and
  Olaf Teschke, editors, {\em Intelligent Computer Mathematics - 10th
  International Conference, {CICM} 2017, Edinburgh, UK, July 17-21, 2017,
  Proceedings}, volume 10383 of {\em Lecture Notes in Computer Science}, pages
  292--302. Springer, 2017.

\bibitem{JakubuvU18}
Jan Jakub\r{u}v and Josef Urban.
\newblock Enhancing {ENIGMA} given clause guidance.
\newblock In Florian Rabe, William~M. Farmer, Grant~O. Passmore, and Abdou
  Youssef, editors, {\em Intelligent Computer Mathematics - 11th International
  Conference, {CICM} 2018, Hagenberg, Austria, August 13-17, 2018,
  Proceedings}, volume 11006 of {\em Lecture Notes in Computer Science}, pages
  118--124. Springer, 2018.

\bibitem{DBLP:journals/aicom/JakubuvU18}
Jan Jakub\r{u}v and Josef Urban.
\newblock Hierarchical invention of theorem proving strategies.
\newblock {\em {AI} Commun.}, 31(3):237--250, 2018.

\bibitem{JakubuvU19}
Jan Jakub\r{u}v and Josef Urban.
\newblock Hammering {Mizar} by learning clause guidance.
\newblock In John Harrison, John O'Leary, and Andrew Tolmach, editors, {\em
  10th International Conference on Interactive Theorem Proving, {ITP} 2019,
  September 9-12, 2019, Portland, OR, {USA}}, volume 141 of {\em LIPIcs}, pages
  34:1--34:8. Schloss Dagstuhl - Leibniz-Zentrum f{\"{u}}r Informatik, 2019.

\bibitem{DBLP:conf/tableaux/Kaliszyk15}
Cezary Kaliszyk.
\newblock Efficient low-level connection tableaux.
\newblock In Hans de~Nivelle, editor, {\em Automated Reasoning with Analytic
  Tableaux and Related Methods - 24th International Conference, {TABLEAUX}
  2015, Wroc{\l}aw, Poland, September 21-24, 2015. Proceedings}, volume 9323 of
  {\em Lecture Notes in Computer Science}, pages 102--111. Springer, 2015.

\bibitem{KaliszykU13b}
Cezary Kaliszyk and Josef Urban.
\newblock {MizAR 40 for Mizar 40}.
\newblock {\em J. Autom. Reasoning}, 55(3):245--256, 2015.

\bibitem{KaliszykUMO18}
Cezary Kaliszyk, Josef Urban, Henryk Michalewski, and Miroslav Ols{\'{a}}k.
\newblock Reinforcement learning of theorem proving.
\newblock In {\em Advances in Neural Information Processing Systems 31: Annual
  Conference on Neural Information Processing Systems 2018, NeurIPS 2018, 3-8
  December 2018, Montr{\'{e}}al, Canada.}, pages 8836--8847, 2018.

\bibitem{LightGBM}
Guolin Ke, Qi~Meng, Thomas Finley, Taifeng Wang, Wei Chen, Weidong Ma, Qiwei
  Ye, and Tie{-}Yan Liu.
\newblock Lightgbm: {A} highly efficient gradient boosting decision tree.
\newblock In {\em {NIPS}}, pages 3146--3154, 2017.

\bibitem{KinyonVV13}
Michael~K. Kinyon, Robert Veroff, and Petr Vojtechovsk{\'{y}}.
\newblock Loops with abelian inner mapping groups: An application of automated
  deduction.
\newblock In Maria~Paola Bonacina and Mark~E. Stickel, editors, {\em Automated
  Reasoning and Mathematics - Essays in Memory of William W. McCune}, volume
  7788 of {\em LNCS}, pages 151--164. Springer, 2013.

\bibitem{Vampire}
Laura Kov{\'a}cs and Andrei Voronkov.
\newblock First-order theorem proving and {V}ampire.
\newblock In Natasha Sharygina and Helmut Veith, editors, {\em CAV}, volume
  8044 of {\em LNCS}, pages 1--35. Springer, 2013.

\bibitem{LoosISK17}
Sarah~M. Loos, Geoffrey Irving, Christian Szegedy, and Cezary Kaliszyk.
\newblock Deep network guided proof search.
\newblock In Thomas Eiter and David Sands, editors, {\em LPAR-21, 21st
  International Conference on Logic for Programming, Artificial Intelligence
  and Reasoning, Maun, Botswana, May 7-12, 2017}, volume~46 of {\em EPiC Series
  in Computing}, pages 85--105. EasyChair, 2017.

\bibitem{McCune92a}
William McCune.
\newblock Experiments with discrimination-tree indexing and path indexing for
  term retrieval.
\newblock {\em J. Autom. Reason.}, 9(2):147--167, 1992.

\bibitem{DBLP:conf/ecai/OlsakKU20}
Miroslav Ols{\'{a}}k, Cezary Kaliszyk, and Josef Urban.
\newblock Property invariant embedding for automated reasoning.
\newblock In Giuseppe~De Giacomo, Alejandro Catal{\'{a}}, Bistra Dilkina,
  Michela Milano, Sen{\'{e}}n Barro, Alberto Bugar{\'{\i}}n, and
  J{\'{e}}r{\^{o}}me Lang, editors, {\em {ECAI} 2020 - 24th European Conference
  on Artificial Intelligence, 29 August-8 September 2020, Santiago de
  Compostela, Spain, August 29 - September 8, 2020 - Including 10th Conference
  on Prestigious Applications of Artificial Intelligence {(PAIS} 2020)}, volume
  325 of {\em Frontiers in Artificial Intelligence and Applications}, pages
  1395--1402. {IOS} Press, 2020.

\bibitem{Overbeek:1974:NCA:321812.321814}
Ross~A. Overbeek.
\newblock A new class of automated theorem-proving algorithms.
\newblock {\em J. ACM}, 21(2):191--200, April 1974.

\bibitem{DBLP:conf/cade/RathsO08}
Thomas Raths and Jens Otten.
\newblock randocop: Randomizing the proof search order in the connection
  calculus.
\newblock In Boris Konev, Renate~A. Schmidt, and Stephan Schulz, editors, {\em
  Proceedings of the First International Workshop on Practical Aspects of
  Automated Reasoning, Sydney, Australia, August 10-11, 2008}, volume 373 of
  {\em {CEUR} Workshop Proceedings}. CEUR-WS.org, 2008.

\bibitem{DBLP:conf/cade/Ruhdorfer020}
Constantin Ruhdorfer and Stephan Schulz.
\newblock Efficient implementation of large-scale watchlists.
\newblock In Pascal Fontaine, Konstantin Korovin, Ilias~S. Kotsireas, Philipp
  R{\"{u}}mmer, and Sophie Tourret, editors, {\em Joint Proceedings of the 7th
  Workshop on Practical Aspects of Automated Reasoning {(PAAR)} and the 5th
  Satisfiability Checking and Symbolic Computation Workshop (SC-Square)
  Workshop, 2020 co-located with the 10th International Joint Conference on
  Automated Reasoning {(IJCAR} 2020), Paris, France, June-July, 2020
  (Virtual)}, volume 2752 of {\em {CEUR} Workshop Proceedings}, pages 120--133.
  CEUR-WS.org, 2020.

\bibitem{SchaferS15}
Simon Sch{\"{a}}fer and Stephan Schulz.
\newblock Breeding theorem proving heuristics with genetic algorithms.
\newblock In Gottlob et~al. \cite{DBLP:conf/gcai/2015}, pages 263--274.

\bibitem{Schulz12}
Stephan Schulz.
\newblock Fingerprint indexing for paramodulation and rewriting.
\newblock In Bernhard Gramlich, Dale Miller, and Uli Sattler, editors, {\em
  Automated Reasoning - 6th International Joint Conference, {IJCAR} 2012,
  Manchester, UK, June 26-29, 2012. Proceedings}, volume 7364 of {\em Lecture
  Notes in Computer Science}, pages 477--483. Springer, 2012.

\bibitem{Schulz13}
Stephan Schulz.
\newblock System description: {E} 1.8.
\newblock In Kenneth~L. McMillan, Aart Middeldorp, and Andrei Voronkov,
  editors, {\em LPAR}, volume 8312 of {\em LNCS}, pages 735--743. Springer,
  2013.

\bibitem{Schulz19}
Stephan Schulz, Simon Cruanes, and Petar Vukmirovic.
\newblock Faster, higher, stronger: {E} 2.3.
\newblock In Pascal Fontaine, editor, {\em Automated Deduction - {CADE} 27 -
  27th International Conference on Automated Deduction, Natal, Brazil, August
  27-30, 2019, Proceedings}, volume 11716 of {\em Lecture Notes in Computer
  Science}, pages 495--507. Springer, 2019.

\bibitem{stickel1989path}
Mark~E Stickel.
\newblock The path-indexing method for indexing terms.
\newblock Technical report, SRI INTERNATIONAL MENLO PARK CA ARTIFICIAL
  INTELLIGENCE CENTER, 1989.

\bibitem{SutcliffeS06}
Geoff Sutcliffe and Christian~B. Suttner.
\newblock The state of {CASC}.
\newblock {\em AI Commun.}, 19(1):35--48, 2006.

\bibitem{SutcliffeU15}
Geoff Sutcliffe and Josef Urban.
\newblock The {CADE-25} automated theorem proving system competition -
  {CASC-25}.
\newblock {\em {AI} Commun.}, 29(3):423--433, 2016.

\bibitem{Urban06}
Josef Urban.
\newblock {MPTP} 0.2: Design, implementation, and initial experiments.
\newblock {\em J. Autom. Reasoning}, 37(1-2):21--43, 2006.

\bibitem{blistr}
Josef Urban.
\newblock {BliStr: The Blind Strategymaker}.
\newblock In Gottlob et~al. \cite{DBLP:conf/gcai/2015}, pages 312--319.

\bibitem{UrbanVS11}
Josef Urban, Ji\v{r}\'{\i} Vysko\v{c}il, and Petr \v{S}t\v{e}p{\'a}nek.
\newblock {MaLeCoP}: Machine learning connection prover.
\newblock In Kai Br{\"u}nnler and George Metcalfe, editors, {\em TABLEAUX},
  volume 6793 of {\em LNCS}, pages 263--277. Springer, 2011.

\bibitem{Veroff96}
Robert Veroff.
\newblock Using hints to increase the effectiveness of an automated reasoning
  program: Case studies.
\newblock {\em J. Autom. Reasoning}, 16(3):223--239, 1996.

\bibitem{Voronkov95}
Andrei Voronkov.
\newblock The anatomy of {Vampire} implementing bottom-up procedures with code
  trees.
\newblock {\em J. Autom. Reason.}, 15(2):237--265, 1995.

\end{thebibliography}

\appendix

\section{Strategy $\S$ used in the Experiments} % Section~\ref{sec:experiments}}
\label{sec:str}

The following E strategy has been used to undertake the experimental evaluation.
% in Section~\ref{sec:experiments}.
The given clause selection strategy
(heuristic) is defined using parameter ``\verb+-H+''.

\begin{verbatim}
--definitional-cnf=24 --split-aggressive --simul-paramod -tKBO6 -c1 -F1
-Ginvfreq -winvfreqrank --forward-context-sr --destructive-er-aggressive
--destructive-er --prefer-initial-clauses -WSelectMaxLComplexAvoidPosPred
-H'(1*ConjectureTermPrefixWeight(DeferSOS,1,3,0.1,5,0,0.1,1,4),
    1*ConjectureTermPrefixWeight(DeferSOS,1,3,0.5,100,0,0.2,0.2,4),
    1*Refinedweight(ConstPrio,4,300,4,4,0.7),
    1*RelevanceLevelWeight2(PreferProcessed,0,1,2,1,1,1,200,200,2.5,
                                                           9999.9,9999.9),
    1*StaggeredWeight(DeferSOS,1),
    1*SymbolTypeweight(DeferSOS,18,7,-2,5,9999.9,2,1.5),
    2*Clauseweight(ConstPrio,20,9999,4),
    2*ConjectureSymbolWeight(DeferSOS,9999,20,50,-1,50,3,3,0.5),
    2*StaggeredWeight(DeferSOS,2))'
\end{verbatim}

\section{Results of Parameter Grid Search}
\label{sec:app}

\begin{table}[tp]
\begin{small}
  \caption{\label{Grid1ss}
    Parameter grid search on a 300 problem development dataset for the combination of % reviewer suggested 300-big -> 300-strong
$\ModelSmallLgb$ and $\ModelSmallGnn$  sorted by performance. $\ModelSmallGnn$ alone solves 140 problems.}
\centering
  \begin{tabular}{lllll}
    \toprule
    Threshold & time &  query &   context & solved \\
    \midrule
 0.03 & 60 & 512 & 1024&151 \\
 0.01 & 60 & 512 & 1024&151 \\
 0.03 & 60 & 1024 & 768&149 \\
 0.03 & 30 & 512 & 1024&149 \\
 0.03 & 30 & 1024 & 768&147 \\
 0.01 & 60 & 1024 & 768&146 \\
 0.01 & 30 & 2048 & 768&146 \\
 0.03 & 30 & 256 & 768& 145\\
 0.01 & 30 & 512 & 1024&145 \\
 0.01 & 30 & 256 & 768& 145\\
 0.01 & 30 & 1024 & 768&145 \\
 0.07 & 30 & 512 & 1024&143 \\
 0.05 & 30 & 256 & 768& 143\\
 0.05 & 60 & 256 & 768& 142\\
 0.05 & 30 & 1024 & 768&142 \\
 0.07 & 30 & 512 & 768& 141\\
 0.03 & 30 & 2048 & 768&141 \\
 0.05 & 30 & 512 & 768 &140 \\
 0.05 & 30 & 512 & 1024&140 \\
 0.07 & 30 & 2048 & 768&139 \\
 0.07 & 30 & 1024 & 768&138 \\
 0.1 & 30 & 512 & 768  &137 \\
 0.1 & 30 & 256 & 768  &137 \\
 0.1 & 30 & 1024 & 768 &137 \\
 0.07 & 30 & 256 & 768& 137\\
 0.05 & 30 & 2048 & 768&137 \\
 0.1 & 30 & 512 & 1024& 136\\
    0.1 & 30 & 2048 & 768& 134\\
       \bottomrule
    \end{tabular}
\quad
 \begin{tabular}{lllll}
    \toprule
    Threshold & time &  query &   context & solved \\
    \midrule
    0.2 & 30 & 512 & 768&  132\\
 0.2 & 30 & 256 & 768&  131\\
 0.3 & 30 & 256 & 768&  130\\
 0.2 & 30 & 512 & 1024& 129\\
 0.2 & 30 & 2048 & 768& 129\\
 0.3 & 30 & 512 & 768 & 127\\
 0.3 & 30 & 512 & 1024& 127\\
 0.3 & 30 & 2048 & 768& 126\\
 0.4 & 30 & 256 & 768 & 121\\
 0.4 & 30 & 512 & 768 & 119\\
 0.5 & 30 & 512 & 768 & 118\\
 0.4 & 30 & 512 & 1024& 118\\
 0.4 & 30 & 2048 & 768& 118\\
 0.5 & 30 & 512 & 1024& 117\\
 0.5 & 30 & 256 & 768 & 114\\
 0.5 & 30 & 2048 & 768& 113\\
 0.6 & 30 & 2048 & 768& 108\\
 0.6 & 30 & 512 & 768 & 106\\
 0.6 & 30 & 512 & 1024& 105\\
 0.6 & 30 & 256 & 768 & 104\\
 0.7 & 30 & 512 & 768 & 103\\
 0.7 & 30 & 512 & 1024& 103\\
 0.7 & 30 & 2048 & 768& 101\\
 0.7 & 30 & 256 & 768 & 100\\
 0.8 & 30 & 512 & 768 & 97 \\
 0.8 & 30 & 512 & 1024& 97 \\
 0.8 & 30 & 2048 & 768& 97 \\
 0.8 & 30 & 256 & 768 & 94 \\
    \bottomrule
  \end{tabular}

% \begin{tabular}{lllll}
%     \toprule
%     Threshold & time &  query &   context & solved \\
%     \midrule
%  0.03 & 60 & 512 & 1024&151 \\
%  0.01 & 60 & 512 & 1024&151 \\
%  0.03 & 60 & 1024 & 768&149 \\
%  0.03 & 30 & 512 & 1024&149 \\
%  0.03 & 30 & 1024 & 768&147 \\
%  0.01 & 60 & 1024 & 768&146 \\
%        \bottomrule
%     \end{tabular}
% \quad
%  \begin{tabular}{lllll}
%     \toprule
%     Threshold & time &  query &   context & solved \\
%     \midrule
% 0.01 & 30 & 2048 & 768&146 \\
%  0.03 & 30 & 256 & 768& 145\\
%  0.01 & 30 & 512 & 1024&145 \\
%  0.01 & 30 & 256 & 768& 145\\
%  0.01 & 30 & 1024 & 768&145 \\
% 0.07 & 30 & 512 & 1024&143 \\
%    \bottomrule
%   \end{tabular}
\end{small}
%\vspace{-2mm}
\end{table}

\begin{table}[tb]
\begin{small}
  \caption{\label{Grid1ls} Parameter grid search on a 300-big development dataset for the combination of
$\ModelLargeLgb$ and $\ModelSmallGnn$ sorted by performance. $\ModelSmallGnn$ alone solves 140 problems.}
\centering
  \begin{tabular}{lllll}
    \toprule
    Threshold & time &  query &   context & solved \\
    \midrule
0.4 & 60 & 2048 & 768      & 164    \\
 0.3 & 30 & 2048 & 768      & 163    \\
 0.4 & 30 & 512 & 1024      & 161    \\
 0.3 & 30 & 512 & 768      &  161   \\
 0.2 & 30 & 2048 & 768     & 161    \\
 0.2 & 30 & 256 & 768      &  161   \\
 0.4 & 30 & 512 & 768      &  160   \\
 0.4 & 30 & 256 & 768      &  160   \\
 0.2 & 30 & 512 & 1024      & 160    \\
 0.4 & 30 & 2048 & 768      & 159    \\
 0.3 & 30 & 512 & 1024      & 158    \\
 0.3 & 30 & 256 & 768      &  158   \\
 0.2 & 30 & 2048 & 768      & 156    \\
 0.1 & 30 & 256 & 768      &  156   \\
 0.5 & 30 & 512 & 1024      & 155    \\
 0.1 & 30 & 512 & 1024      & 155    \\
 0.1 & 30 & 2048 & 768      & 155    \\
 0.5 & 30 & 256 & 768      &  154   \\
 0.2 & 30 & 512 & 768      &  154   \\
 0.5 & 30 & 512 & 768      &  153   \\
 0.5 & 30 & 2048 & 768      & 152    \\
    \bottomrule
  \end{tabular}
\quad
  \begin{tabular}{lllll}
    \toprule
    Threshold & time &  query &   context & solved \\
    \midrule
0.1 & 30 & 512 & 768      &  152   \\
 0.07 & 30 & 512 & 768      & 152    \\
 0.05 & 30 & 512 & 1024    &  152   \\
 0.07 & 30 & 2048 & 768    &  149   \\
 0.05 & 30 & 256 & 768     &  149   \\
 0.05 & 30 & 2048 & 768    &  148   \\
 0.05 & 30 & 512 & 768      & 147    \\
 0.07 & 30 & 512 & 1024    &  146   \\
 0.07 & 30 & 256 & 768      & 146    \\
 0.6 & 30 & 256 & 768      &  144   \\
 0.6 & 30 & 512 & 768      &  143   \\
 0.6 & 30 & 512 & 1024      & 143    \\
 0.6 & 30 & 2048 & 768      & 137    \\
 0.7 & 30 & 256 & 768      &  122   \\
 0.7 & 30 & 512 & 768      &  121   \\
 0.7 & 30 & 512 & 1024      & 121    \\
 0.7 & 30 & 2048 & 768      & 120    \\
 0.8 & 30 & 512 & 768      &  106   \\
 0.8 & 30 & 512 & 1024      & 106    \\
 0.8 & 30 & 256 & 768      &  106   \\
 0.8 & 30 & 2048 & 768      & 103    \\
    \bottomrule
  \end{tabular}
% \begin{tabular}{lllll}
%     \toprule
%     Threshold & time &  query &   context & solved \\
%     \midrule
%  0.4 & 60 & 2048 & 768      & 164    \\
%  0.3 & 30 & 2048 & 768      & 163    \\
%  0.4 & 30 & 512 & 1024      & 161    \\
%  0.3 & 30 & 512 & 768      &  161   \\
%  0.2 & 30 & 2048 & 768     & 161    \\
%  0.2 & 30 & 256 & 768      &  161   \\
%     \bottomrule
%   \end{tabular}
% \quad
%   \begin{tabular}{lllll}
%     \toprule
%     Threshold & time &  query &   context & solved \\
%     \midrule
%  0.4 & 30 & 512 & 768      &  160   \\
%  0.4 & 30 & 256 & 768      &  160   \\
%  0.2 & 30 & 512 & 1024      & 160    \\
%  0.4 & 30 & 2048 & 768      & 159    \\
%  0.3 & 30 & 512 & 1024      & 158    \\
%  0.3 & 30 & 256 & 768      &  158   \\
%     \bottomrule
%   \end{tabular}
\end{small}
%\vspace{-2mm}
\end{table}

\begin{table}[tp]
\begin{small}
\caption{\label{Grid1ll} Parameter grid search on a 300-big development dataset for combinations of $\ModelLargeLgb$ and $\ModelLargeGnn$ sorted by performance. $\ModelLargeGnn$ alone solves 165 problems.}
\centering
  \begin{tabular}{lllll}
    \toprule
    Threshold & time &  query &   context & solved \\
    \midrule
0.1 & 60 & 1024 & 768    &     180          \\
0.2 & 60 & 512 & 1024    &     177          \\
0.1 & 60 & 512 & 1024    &     176          \\
0.1 & 30 & 1024 & 768    &     176          \\
0.2 & 30 & 512 & 1024    &     175          \\
0.1 & 30 & 512 & 1024    &     174          \\
0.1 & 30 & 256 & 768    &      172         \\
0.1 & 30 & 2048 & 768    &     172          \\
0.2 & 30 & 256 & 768    &      171         \\
0.2 & 30 & 1024 & 768    &     171          \\
0.2 & 30 & 2048 & 768    &     170          \\
0.05 & 30 & 1024 & 768    &    170           \\
0.07 & 30 & 1024 & 768    &    168           \\
0.3 & 30 & 256 & 768    &      167         \\
0.3 & 30 & 512 & 1024    &     166          \\
0.3 & 30 & 1024 & 768    &     166          \\
0.4 & 30 & 1024 & 768    &     164          \\
0.3 & 30 & 2048 & 768    &     164          \\
0.4 & 30 & 512 & 1024    &     163          \\
   \bottomrule
  \end{tabular}
    \begin{tabular}{lllll}
    \toprule
    Threshold & time &  query &   context & solved \\
    \midrule
0.4 & 30 & 2048 & 768    &     163          \\
0.4 & 30 & 256 & 768    &      161         \\
0.5 & 30 & 256 & 768    &      158         \\
0.5 & 30 & 512 & 1024    &     156          \\
0.5 & 30 & 1024 & 768    &     155          \\
0.5 & 30 & 2048 & 768    &     151          \\
0.6 & 30 & 256 & 768    &      144         \\
0.6 & 30 & 512 & 1024    &     143          \\
0.6 & 30 & 1024 & 768    &     138          \\
0.6 & 30 & 2048 & 768    &     137          \\
0.7 & 30 & 512 & 1024    &     121          \\
0.7 & 30 & 256 & 768    &      120         \\
0.7 & 30 & 2048 & 768    &     120          \\
0.7 & 30 & 1024 & 768    &     119          \\
0.8 & 30 & 1024 & 768    &     108          \\
0.8 & 30 & 512 & 1024    &     107          \\
0.8 & 30 & 256 & 768    &      107         \\
      0.8 & 30 & 2048 & 768    &     105          \\
      && \\
         \bottomrule
  \end{tabular}

% \begin{tabular}{lllll}
%     \toprule
%     Threshold & time &  query &   context & solved \\
%     \midrule
% 0.1 & 60 & 1024 & 768    &     180          \\
% 0.2 & 60 & 512 & 1024    &     177          \\
% 0.1 & 60 & 512 & 1024    &     176          \\
% 0.1 & 30 & 1024 & 768    &     176          \\
% 0.2 & 30 & 512 & 1024    &     175          \\
% 0.1 & 30 & 512 & 1024    &     174          \\
%    \bottomrule
%   \end{tabular}
%     \begin{tabular}{lllll}
%     \toprule
%     Threshold & time &  query &   context & solved \\
%     \midrule
% 0.1 & 30 & 256 & 768    &      172         \\
% 0.1 & 30 & 2048 & 768    &     172          \\
% 0.2 & 30 & 256 & 768    &      171         \\
% 0.2 & 30 & 1024 & 768    &     171          \\
% 0.2 & 30 & 2048 & 768    &     170          \\
% 0.05 & 30 & 1024 & 768    &    170           \\
%          \bottomrule
%   \end{tabular}
\end{small}
\end{table}
\vspace{-3mm}

\end{document}